\documentclass[11pt,a4paper]{article}
\usepackage[hyperref]{emnlp2018}
\usepackage{times}
\usepackage{latexsym}
\usepackage{amsmath,amsfonts,amsthm}
\usepackage{graphicx}
\usepackage{url}
\usepackage{epsfig} 
\usepackage{multirow}
\usepackage{paralist}
\usepackage{flushend}
\usepackage{epstopdf} 

\renewenvironment{itemize}[1]{\begin{compactitem}#1}{\end{compactitem}}

\newcommand{\argmin}{\operatornamewithlimits{argmin}}

\newcommand{\figref}[1]{Figure \ref{#1}}

\newcommand{\tabref}[1]{Table \ref{#1}}

\aclfinalcopy 

\title{Neural Adaptation Layers for Cross-domain Named Entity Recognition}

\author{
Bill Yuchen Lin\textsuperscript{\dag}\and Wei Lu\textsuperscript{\ddag}\\
\textsuperscript{\dag}University of Southern California\\
\textsuperscript{\ddag}Singapore University of Technology and Design\\
  {\tt yuchen.lin@usc.edu, luwei@sutd.edu.sg} 
}

\date{}

\begin{document} 

\maketitle
\begin{abstract} 
Recent research efforts have shown that neural architectures can be effective in conventional information extraction tasks such as named entity recognition, yielding state-of-the-art results on standard newswire datasets.
However, despite significant resources required for training such models, the performance of a  model trained on one domain typically degrades dramatically when applied to a different domain, yet extracting entities from new emerging domains such as social media can be of significant interest. 
In this paper, we empirically investigate effective methods for conveniently adapting an existing, well-trained neural NER model for a new domain.
Unlike existing approaches, we propose lightweight yet effective methods for performing domain adaptation for neural models.
Specifically, we introduce adaptation layers on top of existing neural architectures, where no re-training using the source domain data is required.
We conduct extensive empirical studies and show that our approach significantly outperforms state-of-the-art methods.\let\thefootnote\relax\footnote{Accepted as a long paper in EMNLP 2018 (Conference on Empirical Methods in Natural Language Processing).} 
\end{abstract}

\section{Introduction}
\label{sec:intro}

Named entity recognition (NER) focuses on extracting named entities in a given text while identifying their underlying semantic types.
%
%
%
Most earlier approaches to NER are based on conventional structured prediction models such as conditional random fields (CRF) \cite{lafferty2001conditional,Sarawagi2004SemiMarkovCR}, relying on hand-crafted features which can be designed based on domain-specific knowledge~\cite{Yang:2012:EOE:2390948.2391100,passos2014lexicon,luo2015joint}. 
Recently, neural architectures have been shown effective in such a task, whereby minimal feature engineering is required \cite{DBLP:conf/naacl/LampleBSKD16,DBLP:conf/acl/MaH16,Peters2017SemisupervisedST,liu2017empower}. 
%
\begin{figure}[t]
	\centering 
	\epsfig{file=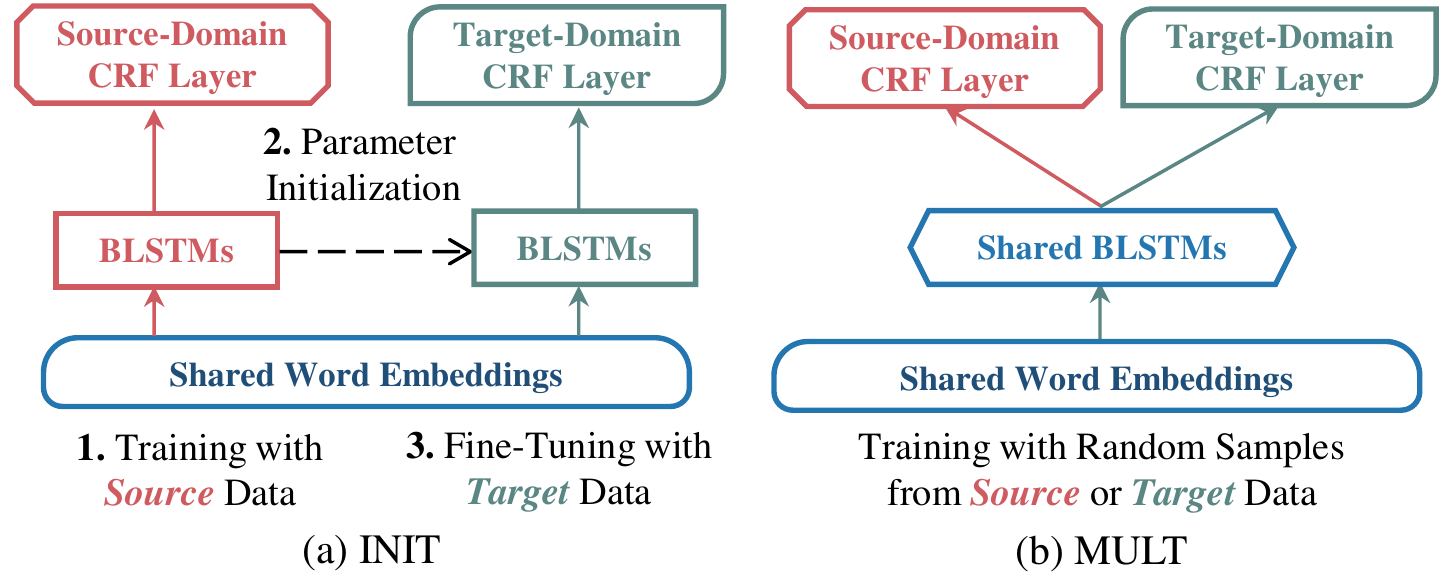, width=1.0\columnwidth}
	\caption{{Two existing adaptation approaches for NER.}}
	\label{fig:initmult} 
	\vspace{-10pt}
\end{figure}
Domain adaptation, as a special case for transfer learning, aims to exploit the abundant data of well-studied source domains to improve the performance in {target} domains of interest~\cite{DBLP:journals/tkde/PanY10,weiss2016survey}. 
There is a growing interest in investigating the transferability of neural  models for NLP.
Two notable approaches, namely INIT (parameter initialization) and MULT (multitask learning), have been proposed for studying the transferrability of neural networks under tasks such as sentence (pair) classification \cite{Mou2016HowTA} and sequence labeling \cite{Yang2016TransferLF}.

\begin{figure*}[t!!]
\centering
\epsfig{file=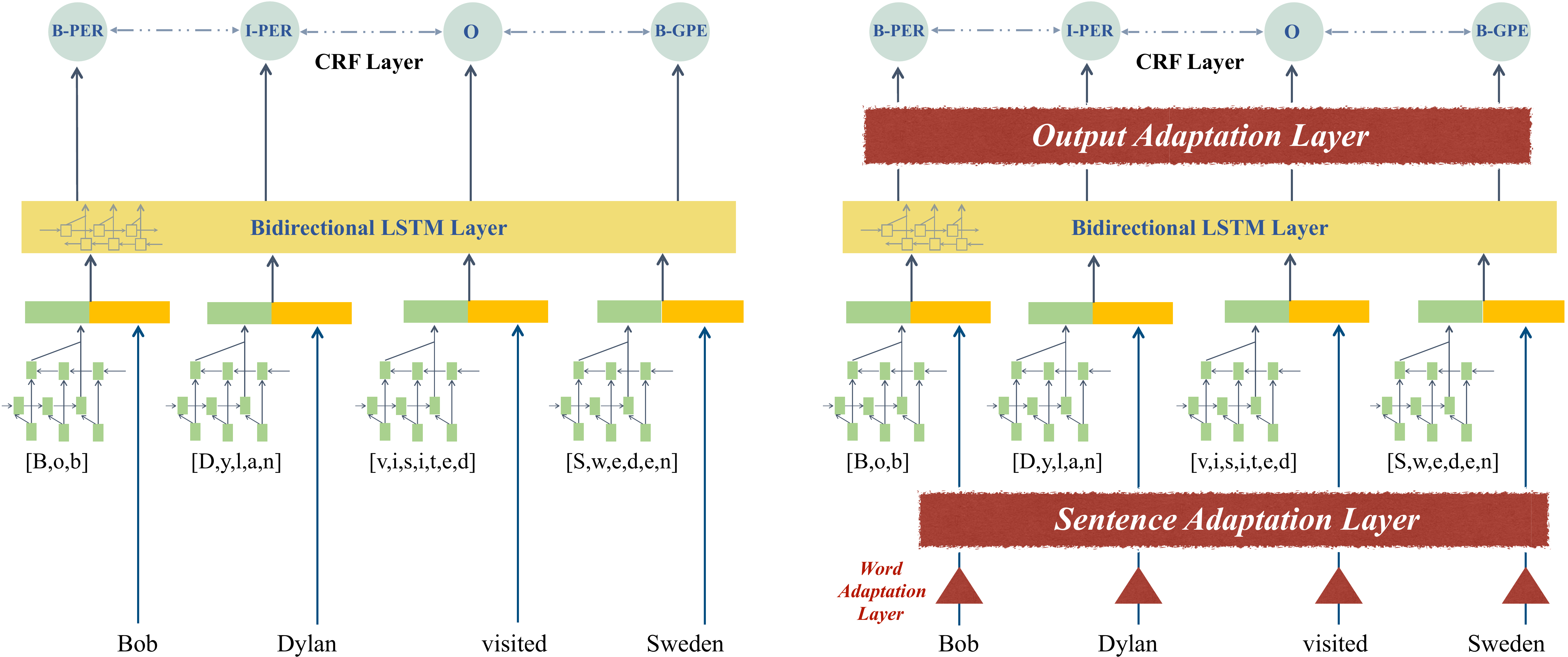, width=2.1\columnwidth} 
\caption{{Our base model (left) and target  model (right) where we insert new adaptation layers (highlighted in red).}}
\label{fig:ourmodel}
\end{figure*}

%

The INIT method first trains a  model using labeled data from the source domain; next, it initializes a target model with the learned parameters; 
finally, it fine-tunes the initialized target model using labeled data from the target domain.
The MULT method, on the other hand,  simultaneously trains two models using both source and target data respectively, where some parameters are shared across the two models during the learning process.
Figure \ref{fig:initmult} illustrates the two approaches based on the BLSTM-CRF (bidirectional LSTM augmented with a CRF layer) architecture for NER.
While such approaches make intuitive senses, they also come with some limitations.

First, these methods utilize shared domain-general word embeddings when performing learning from both source and target domains.
This essentially assumes there is no domain shift of input feature spaces. 
However, cases when the two domains are distinct (words may contain different semantics across two domains), we believe such an assumption can be weak.

Second, existing approaches such as INIT
directly augment the LSTM layer with a new output CRF layer when learning models for the target domain.
One basic assumption involved here is that the model would be able to re-construct a new CRF layer that can capture not only the variation of the input features (or hidden states outputted from LSTM) to the final CRF layer across two domains, but also the structural dependencies between output nodes in the target output space.
We believe this overly restrictive assumption may prevent the model from capturing rich, complex cross-domain information due to the inherent linear nature of the CRF model.

Third, most methods involving cross-domain embedding often require highly time-consuming retraining word embeddings on domain-specific corpora. 
This makes it less realistic in scenarios where source corpora are huge or even inaccessible.
Also, MULT-based methods need retraining on the source-domain data for different target domains.
We think this disadvantage of existing methods hinders the neural domain adaptation methods for NER to be practical in real scenarios.

In this work, we propose solutions to address the above-mentioned issues.
Specifically, we address the first issue at both the word and sentence level by introducing a {\it word adaptation layer} and {\em a sentence adaptation layer} respectively, bridging the gap between the two input spaces.
Similarly, an {\em output adaptation layer} is also introduced between the LSTM and the final CRF layer, capturing the variations in the two output spaces.
Furthermore, we introduce a single hyper-parameter that controls how much information we would like to preserve from the model trained on the source domain.
These approaches are lightweight, without requiring re-training on data from the source domain.
We show through extensive empirical analysis as well as ablation study that our proposed approach can significantly improve the performance of cross-domain NER over existing transfer approaches.

\section{Base Model}\label{sec:basemodel}
We briefly introduce the BLSTM-CRF architecture for NER, which serves as our {\em base model} throughout this paper.
Our base model is the combination of two recently proposed popular works for named entity recognition by~\citet{DBLP:conf/naacl/LampleBSKD16} and \citet{DBLP:conf/acl/MaH16}.
{Figure~\ref{fig:ourmodel} illustrates the BLSTM-CRF architecture.}

Following~\citet{DBLP:conf/naacl/LampleBSKD16}, we develop the comprehensive word representations by concatenating pre-trained word embeddings and character-level word representations, which are constructed by running a BLSTM over sequences of character embeddings.
The middle BLSTM layer takes a sequence of comprehensive word representations and produces a sequence of hidden states, representing the contextual information of each token.
Finally, following \citet{DBLP:conf/acl/MaH16}, we build the final CRF layer by utilizing potential functions describing local structures to define the conditional probabilities of complete predictions for the given input sentence.

This architecture is selected as our base model due to its generality and representativeness. {We note that several recently proposed models~\cite{Peters2017SemisupervisedST,liu2017empower} are built based on it.}
As our focus is on how to better transfer such architectures for NER, we include further discussions of the model and training details in our supplementary material.

\section{Our Approach}\label{sec:approach}

We first introduce our proposed three adaptation layers and  describe the overall learning process.

\subsection{Word Adaptation Layer}\label{sec:cdemb}

Most existing transfer approaches use the same domain-general word embeddings for training both source and target models.
Assuming that there is {little} domain shift of input feature spaces, they simplify the problem as \textit{homogeneous transfer learning}~{\cite{weiss2016survey}}.
However, this simplified assumption becomes weak when two domains have apparently different language styles and involve {considerable} domain-specific terms that may not share the same semantics across the domains; {for example, the term ``{\em cell}'' has different meaning in biomedical articles and product reviews.}
Furthermore, some important domain-specific words may not be present in the vocabulary of domain-general embeddings.
As a result, we have to regard such words as out-of-vacabulary (OOV) words, which may be harmful to the transfer learning process.

{\citet{stenetorp2012size} show that domain-specific word embeddings tend to perform better when used in supervised learning tasks.\footnote{{We also confirm this claim with experiments presented in supplementary materials.}}
}
However, maintaining such an improvement in the transfer learning process is very challenging.
This is because two domain-specific embeddings are trained separately  on source and target datasets, and therefore the two embedding spaces are \textit{heterogeneous}.
Thus, model parameters trained in each model are heterogeneous as well, which hinders the transfer process.
How can we address such challenges while maintaining the improvement by using domain-specific embeddings?

We address this issue by developing a word adaptation layer, bridging the gap between the source and target embedding spaces, so that both input features and model parameters become homogeneous across domains.
Popular pre-trained word embeddings are usually trained on very large corpora, and thus methods requiring re-training them can be extremely costly and time-consuming.
We propose a straightforward{, lightweight} yet effective method to construct the word adaptation layer that projects the learned embeddings from the target embedding space into the source space.
This method only requires some corpus-level statistics from the datasets (used for learning embeddings) to build the \textit{pivot lexicon} for constructing the adaptation layer.

\subsubsection{Pivot Lexicon}


A \textit{pivot word pair} is denoted as $(w_s,w_t)$, where $w_s \in \mathcal{X}_S$ and $w_t \in \mathcal{X}_T$. 
Here, $\mathcal{X}_S$ and $\mathcal{X}_T$ are source and target vocabularies.
A \textit{pivot lexicon} $\mathcal{P}$ is a set of such word pairs.
    
To construct a pivot lexicon,
first, motivated by {\citet{Tan2015LexicalCB}}, we define $\mathcal{P}_1$, which consists of the ordinary words that have high relative frequency in both source and target domain corpora:
{ $\mathcal{P}_1=\{ (w_s, w_t)| w_s=w_t , f(w_s)\ge\phi_s, f(w_t) \ge \phi_t \} $}, where $f(w)$ is the frequency function that returns the number of occorrence of the word $w$ in the dataset, and $\phi_s$ and $\phi_t$ are word frequency thresholds\footnote{A simple way of setting them is to choose the frequency of the $k$-th word in the word lists sorted by frequencies.
 }.
Optionally, we can utilize a customized word-pair  list $\mathcal{P}_2$, which gives mappings between domain-specific words across domains, such as normalization lexicons {\cite{Han2011LexicalNO, Liu2012ABN}}.
The final lexicon is thus defined as $\mathcal{P}= \mathcal{P}_1  \cup \mathcal{P}_2$.
    
\subsubsection{Projection Learning}

Mathematically, given the pre-trained domain-specific word embeddings $\mathbf{V_S}$ and $\mathbf{V_T}$ as well as a pivot lexicon $\mathcal{P}$, we would like to learn a linear projection transforming word vectors from $\mathbf{V_T}$ into $\mathbf{V_S}$. 
This idea is based on a bilingual word embedding model~\cite{Mikolov2013ExploitingSA}, but we adapt it to this domain adaptation task.

We first construct two matrices $\mathbf{V^*_S}$ and $\mathbf{V^*_T}$, where the $i$-${\text{th}}$ rows of these two matrices are the vector representations for the words in the $i$-${\text{th}}$ entry of $\mathcal{P}$: $\mathcal{P}^{(i)} = (w^{i}_s, w^{i}_t)$.
We use $\mathbf{V^{*i}_{S}}$ to denote the vector representation of the word $w^i_{s}$, and similarly for $\mathbf{V^{*i}_{T}}$ and $w^i_{t}$. 
    
Next, we learn a transformation matrix $\mathbf{Z}$ minimizing the distances between $\mathbf{V^{*}_{S}}$ and $\mathbf{V^{*}_{T}} \mathbf Z$  with the following loss function, where $c_i$ is the confidence coefficient for the entry $\mathcal{P}^{(i)}$, highlighting the significance of the entry:

{\footnotesize $$\argmin_{\mathbf Z} \sum_{i=1}^{|\mathcal{P}|} c_i ~ 
\bigg|\bigg| {\mathbf{V^{*i}_{T}}} \mathbf Z - {\mathbf{V^{*i}_{S}}} \bigg|\bigg|^2
,$$}

We use normalized frequency ($\bar{f}$) and {\color{black}S{\o}rensen-Dice} coefficient \cite{sorensen1948method} to describe the significance of each word pair:

    {\footnotesize $$\bar{f}(w_{s})=\frac{{f}(w_{s})}{\max_{w' \in \mathcal{X}_{S}}{f}(w')},~~\bar{f}(w_{t})=\frac{{f}(w_{t})}{\max_{w' \in \mathcal{X}_{T}}{f}(w')} $$
     $$ c_i = \frac{2 \cdot \bar{f}(w_s^i) \cdot \bar{f}(w_t^i)  }{\bar{f}(w_s^i) + \bar{f}(w_t^i) } $$}
\vspace{-10pt}

{The intuition behind this scoring method is that a word pair is more important when both words have high relative frequency in both domains. This is because such words are likely more domain-independent, conveying identical or similar semantics across these two different domains.}

Now, the matrix $\mathbf{Z}$ exactly gives the weights to the word adaptation layer, which takes in the target domain word embeddings and returns the transformed embeddings in the new space.
We learn $\mathbf Z$ with stochastic gradient descent.
After learning, the projected new embeddings would be ${\mathbf{V_{T}}} \mathbf Z$, which would be used in the subsequent steps as illustrated in Figure \ref{fig:ourmodel} and Figure~\ref{fig:transfer}.
With such a word-level input-space transformation, the parameters of the pre-trained source models based on $\mathbf{V_{S}}$ can still be relevant, which can be used in subsequent steps.

We would like to highlight that, unlike many previous approaches to learning cross-domain word embeddings \cite{Bollegala2015UnsupervisedCW,yang2017}, the learning of our word adaptation layer involves no modifications to the source-domain embedding spaces.
It also requires no re-training of the embeddings based on the target-domain data.
Such a distinctive advantage of our approach comes with some important practical implications: it essentially enables the transfer learning process to work directly on top of a well-trained model by performing  adaptation without involving significant re-training efforts. {For example, the existing model could be one that has already gone through extensive training, tuning and testing for months based on large datasets with embeddings learned from a particular domain (which may be different from the target domain).}

\subsection{Sentence Adaptation Layer}\label{sec:inputadpt}

The word adaptation layer serves as a way to bridge the gap of heterogeneous input spaces, but it does so only at the word level and is context independent.
We can still take a step further to address the input space mismatch issue at the sentence level, capturing the contextual information in the learning process of such a mapping based on labeled data from the target domain.
To this end, we augment a BLSTM layer right after the word adaptation layer (see Figure \ref{fig:ourmodel}), and we name it the {\em sentence adaptation layer}.

This layer pre-encodes the sequence of projected word embeddings for each target instance, before they serve as inputs to the LSTM layer inside the base model.
The hidden states for each word generated from this layer can be seen as the further transformed word embeddings capturing target-domain specific contextual information, where such a further transformation is learned in a supervised manner based on target-domain annotations.
We also believe that with such a sentence adaptation layer, the OOV issue mentioned above may also be partially alleviated.
This is because without such a layer, OOV words would all be mapped to a single fixed vector representation -- which is not desirable;
whereas with such a sentence adaptation layer, 
each OOV word would be assigned their ``transformed'' embeddings based on its respective contextual information.

\subsection{Output Adaptation Layer}\label{sec:outputadpt}

We focus on the problem of performing domain adaptation for NER under a general setup, where we assume the set of output labels for the source and target domains could be different. Due to the heterogeneousness of output spaces, we have to reconstruct the final CRF layer in the target models.

However, we believe solely doing this may not be enough to address the {\em labeling difference} problem as highlighted in~\cite{jiang2008domain} as the two output spaces may be very different.
For example, in the sentence ``\textit{\textbf{Taylor} released her new \textbf{songs}}'', ``{\em Taylor}'' should be labeled as ``\textsc{music-artist}'' instead of ``\textsc{person}'' in some social media NER datasets;
this suggests \textit{re-classifying} with contextual information is necessary.
In another example, where we have a tweet ``\textit{so...\textbf{\#kktny} in 30 mins?}'';
here we should  recognize ``{\em \#kktny}'' as a \textsc{creative-work} entity, but there is little similar instances in newswire data, indicating that context-aware \textit{re-recognition} is also needed.
    
How can we perform  re-classification and re-recognition with contextual information in the target model?
We answer this question by inserting a BLSTM {\em output adaptation layer} in the base model, right before the final CRF layer, to capture variations in outputs with contextual information.  
This output adaption layer takes the output hidden states from the BLSTM layer from the base model as its inputs, producing a sequence of new hidden states for the re-constructed CRF layer. 
Without this layer, the learning process directly updates the pre-trained parameters of the base model, which may lead to loss of knowledge that can be transferred from the source domain.

\begin{figure}[t!]
\centering
\epsfig{file=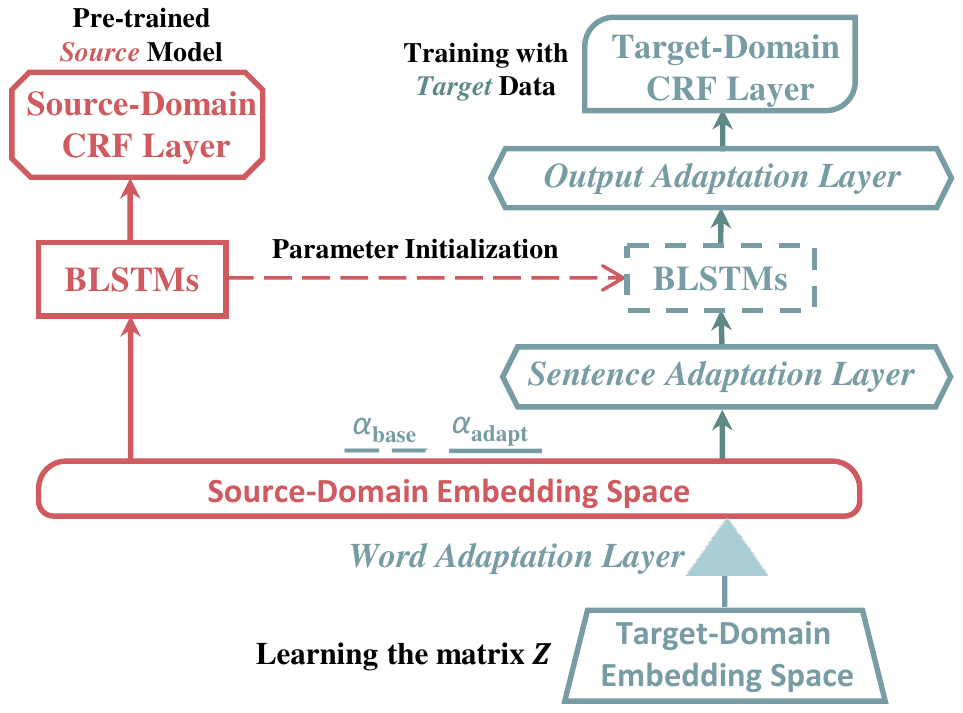, width=1\columnwidth}
\caption{{Overview of our transfer learning process.}}
\label{fig:transfer}
\end{figure}

\subsection{Overall Learning Process}

Figure \ref{fig:transfer} depicts our overall  learning process.
We initialize the base model with the parameters from the pre-trained source model, and tune the weights for all layers --- including layers from the base model, {sentence and output} adaptation layers, and the CRF layer.
We use different learning rates when updating the weights in different layers using {Adam~\cite{kingma2014adam}}. {In all our experiments, we fixed the weights to $\mathbf{Z}$ for the word adaptation layers to avoid over-fitting}. This allows us to preserve the knowledge learned from the source domain while effectively leveraging  the limited training data from the target domain.

Similar to ~\citet{Yang2016TransferLF}, who utilizes a hyper-parameter $\lambda$ for controlling the transferability,
we also introduce a 
hyper-parameter $\psi$ that serves a similar purpose --- it captures the relation between the learning rate used for the base model ($\alpha_\text{base}$) and the learning rate used for the adaptation layers plus the final CRF layer
$\alpha_\text{base} = \psi \cdot \alpha_\text{adapt}$.

If $\psi=0$, we fix the learned parameters (from source domain) of the base model completely (Ours-Frozen).
If $\psi=1$, we treat all the layers equally (Ours-FineTune).

\section{Experimental Setup}\label{sec:eval}
In this section, we present the setup of our experiments. 
We show our choice for source and target domains, resources for embeddings, the datasets for evaluation and finally the baseline methods.

\subsection{Source and Target Domains} 

We evaluate our approach with the setting that the source domain is \textit{newswire} and the target domain is \textit{social media}.
We designed this experimental setup based on the following considerations:

\begin{itemize}
        \item \textit{Challenging}: {Newswire} is a well-studied domain for NER and existing neural models perform very well (around 90.0 F1-score~\cite{DBLP:conf/acl/MaH16}). 
        However, the performance drop dramatically in social media data (around 60.0 F-score~\cite{Strauss2016ResultsOT}).
        \item \textit{Important}: Social media is a rich soil for text mining~\cite{Petrovic2010StreamingFS,Rosenthal2015ICA,Wang2015ThatsSA}, and NER is of significant importance for other information extraction tasks in social media~{\cite{ritter2011,Peng2016ImprovingNE, Chou2016BoostedWN}}.
        \item \textit{Representative}: The noisy nature of user generated content as well as emerging entities with novel surface forms make the domain shift very salient~\cite{Finin2010AnnotatingNE,DBLP:conf/aclnut/2016}.
\end{itemize}
    
{Nevertheless, the techniques developed in this paper are domain independent and thus can be used for other learning tasks across any two domains so long as we have the necessary resources.}
    
\subsection{Resources for Cross-domain Embeddings}

{We utilizes GloVe~\cite{pennington2014glove} to train domain-specific and domain-general word embeddings from different corpora, denoted as follows:  1) {\textit{source\_emb}}, which is trained on the newswire domain corpus (\textit{NewYorkTimes} and \textit{DailyMail} articles); 2)  {\textit{target\_emb}}, which is trained on a social media corpus (Archive Team's Twitter stream grab\footnote{\scriptsize{\url{https://archive.org/details/twitterstream}}});  3) {\textit{general\_emb}}, which is pre-trained on CommonCrawl containing both formal and user-generated content}.\footnote{\scriptsize \url{https://nlp.stanford.edu/projects/glove/}}
{We 
obtain the intersection of the top 5K words from source and target vocabularies sorted by frequency to build $\mathcal{P}_1$.
	For $\mathcal{P}_2$, we utilize an existing twitter normalization lexicon containing 3,802 word pairs\cite{Liu2012ABN}.} More details are in supplemental material.

\subsection{NER Datasets for Evaluation} 

\begin{table}[]
	\small
	\centering 
	\begin{tabular}{c|c|c|c|c}
		&{ \textsc{Co} }&{ \textsc{On}} & {\textsc{Ri}} & {\textsc{Wn}}  \\ \hline 
		\#train\_token    &  204,567   &       848,220       &      37,098      &    46,469       \\
		\#dev\_token      &      51,578     &       144,319    &       4,461     &     16,261      \\ 
		\#test\_token     &      46,666     &        49,235      &       4,730   &      61,908     \\ \hline  
		\#train\_sent. &      14,987     &    33,908          &       1,915     &  2,394 \\ 
		\#dev\_sent.   &      3,466     &       5,771       &           239 &     1,000      \\ 
		\#test\_sent.  &     3,684     &        1,898      &         240   &      3,856     \\ \hline  
		\#entity\_type   &      4     &      11        &     10       &    10    \\  
	\end{tabular}
	\caption{{Statistics of the NER datasets.} \label{nerstat}}
\end{table} 
   
For the source newswire domain, we use the following two datasets: 
OntoNotes-nw -- the newswire section of {O}ntoNotes 5.0 release dataset (\textsc{On})~{\cite{weischedel2013ontonotes}} that is publicly available\footnote{{\scriptsize{\url{https://catalog.ldc.upenn.edu/ldc2013t19}}}}, as well as the {C}oNLL03 NER dataset (\textsc{Co})~{\cite{Sang2003IntroductionTT}}.
For the first dataset, 
we randomly split the dataset into three sets: 80\% for training, 15\% for development and 5\% for testing.
For the second dataset, we follow their provided standard train-dev-test split.
For the target domain, we consider the following two datasets: {R}itter11 (\textsc{Ri})~\cite{Ritter2011NamedER} and {W}NUT16 (\textsc{Wn})~\cite{Strauss2016ResultsOT}, both of which are publicly available. 
The statistics of the four datasets we used in the paper are shown in~\tabref{nerstat}.  
 
    
\subsection{Baseline Transfer Approaches}\label{sec:simapproach}

We present the baseline approaches, which were originally investigated by \citet{Mou2016HowTA}.
\citet{lee2017transfer} explored the INIT method for NER, while \citet{Yang2016TransferLF} extended the MULT method for sequence labeling.
    
\textbf{INIT}: 
We first train a source model $\mathcal{M}_S$ using the source-domain training data $\mathcal{D}_S$. 
Next, we construct a target model $\mathcal{M}_T$ and reconstruct the final CRF layer to address the issue of different output spaces.
We use the learned parameters of $\mathcal{M}_S$ to initialize $\mathcal{M}_T$ excluding the CRF layer.
Finally, {INIT-FineTune }continues training   $\mathcal{M}_T$ with the target-domain training data $\mathcal{D}_T$, while {INIT-Frozen} instead only updates the parameters of the newly constructed CRF layer.


\textbf{MULT}: Multi-task learning based transfer method simultaneously trains $\mathcal{M}_S$ and $\mathcal{M}_T$ using $\mathcal{D}_S$ and $\mathcal{D}_T$. 
The parameters of $\mathcal{M}_S$ and $\mathcal{M}_T$ excluding their CRF layers are shared during the training process.
Both \citet{Mou2016HowTA} and \citet{Yang2016TransferLF} follow \citet{Collobert2008AUA} and use a hyper-parameter $\lambda$ as the probability of choosing an instance from $\mathcal{D}_S$ instead of $\mathcal{D}_T$ to optimize the model parameters.
By selecting the hyper-parameter $\lambda$, the multi-task learning process tends to perform better in target domains.
Note that this method needs re-training of the source model with $\mathcal{D}_S$ every time we would like to build a new target model, which can be time-consuming especially when $\mathcal{D}_S$ is large.

\textbf{MULT+INIT}: This is a combination of the above two methods. 
We first use INIT to initialize the target model.
Afterwards, we train the two models as what MULT does.


\section{Results and Discussion}

\subsection{Main Results}

We primarily focus on the discussion of experiments with a particular setup where $\mathcal{D}_S$ is set to OntoNotes-nw and $\mathcal{D}_T$ is Ritter11.
{In the experiments, ``{in-domain}'' means we only use $\mathcal{D}_T$ to train our base model without any transfer from the source domain.
``$\Delta$'' represents the amount of improvement we can obtain (in terms of $F$ measure) using transfer learning over ``in-domain'' for each transfer method.
The hyper-parameters $\psi$ and $\lambda$ are tuned from $\{0.1, 0.2, ..., 1.0\}$ on the development set,
and we show the results based on the developed hyper-parameters.}

We first conduct the first set of experiments to evaluate 
performance of different transfer methods under the assumption of homogeneous input spaces. 
Thus, we utilize the same word embeddings (general\_emb) for training both source and target models. {Consequently we remove the word adaptation layer (cd\_emb) in our approach under this setting.}
The results are listed in Table \ref{tab:E1}.
As we can observe, the INIT-Frozen method leads to a slight ``negative transfer'', {which is also reported in the experiments of \citet{Mou2016HowTA}}. 
This indicates that solely updating the parameters of the final CRF layer is not enough for performing re-classification and re-recognition of the named entities for the new target domain.
The INIT-FineTune method yields better results for it also updates the parameters of the middle LSTM layers in the base model to mitigate the heterogeneousness.
The MULT and MULT+INIT methods yield higher results, partly due to the fact that they can better control the amount of transfer through tuning the hyper-parameter.
Our proposed transfer approach outperforms all these baseline approaches. It not only controls the amount of transfer across the two domains but also explicitly captures variations in the input and output spaces when there is significant domain shift.

\begin{table}[t!]
\small
\centering
\begin{tabular}{c|c|c|c|c}
Settings & $P$ & $R$ & $F$ & $\Delta$\\ 
\hline 
\underline{in-domain}                        &   72.73       &       56.14    &   {\underline{63.37}} &-     \\
INIT-Frozen                          &   72.61  &       56.11    &   63.30  & -0.07 \\ 
INIT-FineTune                          &   73.13  &       56.55     &   63.78  & +0.41 \\ 
MULT
&   74.07    &     57.51    &  {64.75} & +1.38  \\ 
MULT+INIT 
&   74.12    &     57.57    &  {64.81} & +1.44  \\
\hline
Ours ({\em w/o word adapt})
&   74.87    &    57.95    &   \textbf{65.33}  & \textbf{+1.96}\\ 
\end{tabular}
\caption{{Comparisons of different methods for \textbf{homogeneous} input spaces. ($\mathcal{D}_S$ = \textsc{On}, $\mathcal{D}_T$ = \textsc{Ri})}}
\label{tab:E1}
\end{table}  

We use the second set of experiments to understand the effectiveness of each method when dealing with heterogeneous input spaces.
We use source\_emb for training source models and target\_emb for learning the target models.
From the results shown in Table~\ref{tab:E2}, we can find that all the baseline methods degrade when the two word embeddings used for training source models and learning target models are different from each other.
The heterogeneousness of input feature space hinders them to better use the information from source domains.
However, with the help of word   and sentence adaptation layers, our method achieves better results.
The experiment on learning without the word adaptation layer also confirms the importance of such a layer.\footnote{We use the approximate randomization test~\cite{Yeh2000MoreAT} for statistical significance of the difference
	between ``Ours'' and ``MULT+INIT''. Our improvements are statistically
	significant with a $p$-value of 0.0033.  }

Our results are also comparable to the results when the cross-lingual embedding method of~\citet{yang2017} is used instead of the word adaptation layer.
However, as we mentioned earlier, their method requires re-training the embeddings from target domain, and is more expensive.

\subsection{Ablation Test}

To investigate the effectiveness of each component of our method, we conduct  ablation test based on our full model ($F$=66.40) reported in Table \ref{tab:E2}. 
We use $\Delta$ to denote the differences of the performance between each setting and our model.
The results of ablation test are shown in Table \ref{tab:E3}. 
We first set $\psi$ to 0 and 1 respectively to investigate the two special variant (Ours-Frozen, Ours-FineTune) of our method.
We find they both perform worse than using the optimal $\psi$ (0.6). 

One natural concern is whether our performance gain is truly caused by the effective approach for cross-domain transfer learning, or is simply because we use a new architecture with more layers (that is built on top of the base model) for learning the target model.
{To understand this, we carry out an experiment named ``w/o transfer'' by setting $\psi$ to 1, where we randomly initialize the parameters of the middle BLSTMs in the target model instead of using source model parameters.}
Results show that such a model does not perform well, confirming the effectiveness of  transfer learning with our proposed adaptation layers.
Results also confirm the importance of all the three adaptation layers that we introduced.
Learning the confidence scores ($c_i$) and employing the optional $\mathcal{P}_2$ are also helpful but they appear to be playing less significant roles.


\begin{table}[t]
	\small 
	\centering
	\begin{tabular}{c|c|c|c|c}
		Settings & $P$ & $R$ & $F$ & $\Delta$\\ 
		\hline 
		\underline{in-domain}                        &   72.51       &       57.11    &   {\underline{63.90}} &-     \\
		INIT-Frozen                          &    72.65  &       55.25     &    62.77  & -1.13  \\  
		INIT-FineTune                          &   72.83  &       56.73     &    63.78  & -0.12  \\   
		MULT     &   73.11    &     57.35    &  {64.28}  & +0.38  \\  
		MULT+INIT     &   73.13    &     57.31    &  {64.26}  & +0.36  \\  
		\hline
		Ours         &   75.87     &     59.03    &  {66.40}  & {+2.50}     \\ 
		w/ \citet{yang2017}        &   76.12    &     59.10   &  {66.53}  & {+2.63}     \\ 
		{\em w/o word adapt. layer}         &   73.29     &     57.61    &  {64.51}  & {+0.61}     \\

	\end{tabular}
	\caption{{Comparisons of different methods for \textbf{heterogeneous} input spaces. ($\mathcal{D}_S$ = \textsc{On}, $\mathcal{D}_T$ = \textsc{Ri})}}
	\label{tab:E2}
\end{table}  

\subsection{Additional Experiments}
As shown in Table~\ref{tab:outofdomain2}, we conduct some additional experiments to investigate the significance of our improvements on different source-target domains,
and whether the improvement is simply because of the increased model expressiveness due to a larger number of parameters.\footnote{In this table, I-200/300 and M-200/300 represent the best performance from \{INIT-Frozen, INIT-FineTune\} and \{MULT, MULT+INIT\} respectively after tuning; ``in-do.'' here is the best score of our base model without transfer.}


We first set the hidden dimension to be the same as the dimension of source-domain word embeddings for the sentence adaptation layer, which is 200 (I/M-200).
The dimension used for the output adaptation layer is just half of that of the base BLSTM model. 
Overall, our model roughly involves 117.3\% more parameters than the base model.\footnote{Base Model: $(25\times 50+50^2) + (250\times 200+200^2) = 93,750$; Ours: $(25\times 50+50^2) + (200\times 200+200^2) + (250\times 200+200^2) + (200\times 100+100^2) = 203,750$.}
To understand the effect of a larger parameter size, we further experimented with hidden unit size as $300$ (I/M-300), leading to a parameter size of $213,750$ that is comparable to ``Ours'' ($203,750$). As we can observe, our approach  outperforms these approaches consistently in the four settings.
More experiments with other settings can be found in the supplementary material.

\begin{table}[]
\small
\centering
\begin{tabular}{c|l|c|c}
\multicolumn{2}{c|}{Settings} & $F$ & $\Delta$\\ 
\hline 
$\psi$=0.0 &    Ours-Frozen       &    63.95   & -2.45 \\
$\psi$=1.0    & Ours-FineTune   &  63.40    & -3.00 \\ 
{$\psi$=1.0} &{w/o transfer}        &   63.26   & -3.14 \\ 
{$\psi$=0.6} &{w/o using confidence $c_i$}    &  {66.04}  & -0.36 \\ 
{$\psi$=0.6} &{w/o using $\mathcal{P}_2$}    &  {66.11}  & -0.29 \\ 
{$\psi$=0.6} &{w/o word adapt. layer}        &     64.51 & -1.89 \\ 
{$\psi$=0.6}& {w/o sentence adapt. layer}       &   65.25  & -1.15 \\
{$\psi$=0.6} &{w/o output adapt. layer}    &   64.84   & -1.56 \\ 
\end{tabular}
\caption{{{Comparing different settings of our method.
}}
\label{tab:E3}
}
\end{table}

\begin{table}[t!]
\scriptsize
\centering
\begin{tabular}{c|c|c|c|c|c|c}
$\mathcal{D}_S, \mathcal{D}_T$ & in-do. &   I-200 & M-200 & I-300 & M-300 & Ours \\ 
\hline
{\scriptsize (\textsc{On}, \textsc{Ri})}&         \multirow{2}{*}{63.37} & +0.41    &  +1.44  & +0.43 & +1.48  &  +3.03     \\  
{\scriptsize (\textsc{Co}, \textsc{Ri})} &           &       +0.23 &  +0.81 & +0.22  & + 0.88& +1.86\\   \hline 
{\scriptsize (\textsc{On}, \textsc{Wn})}&                \multirow{2}{*}{51.03}        &     +0.89    &  +1.72 & +0.88 & +1.77 & +3.16    \\  
{\scriptsize (\textsc{Co}, \textsc{Wn})} &         &     +0.69     &  +1.04 & +0.71 & +1.13 & +2.38 \\ 
%
\end{tabular}
\caption{{ Results of transfer learning methods on different datasets with different number of LSTM units in the base model. (I: INIT; M: MULT; 200/300: number of LSTM units).}  }
\label{tab:outofdomain2}
\end{table}

\subsection{Effect of Target-Domain Data Size} 

\begin{figure}[t!]
\centering
\resizebox{1\columnwidth}{!}{\input{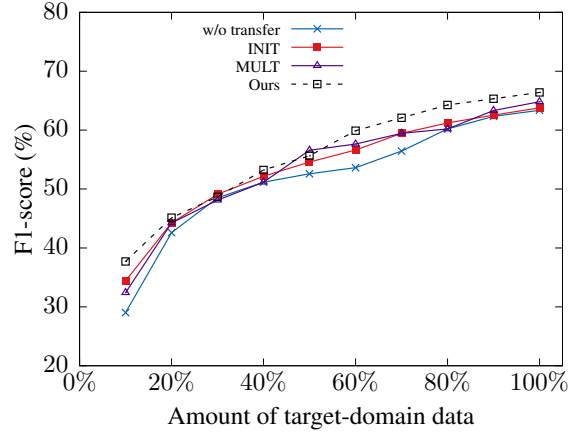}}
\caption{{\footnotesize F1-score vs.  amount of target-domain data.\vspace{-15pt}}}
\label{fig:diffrate}
\end{figure}

To assess the effectiveness of our approach when we have different amounts of training data from the target domain, we conduct additional experiments by gradually increasing the amount of the target training data from 10\% to 100\%.
{We again use the OntoNotes-nw and Ritter11 as $\mathcal{D}_S$ and $\mathcal{D}_T$, respectively.
Results are shown in Figure \ref{fig:diffrate}.
Experiments for INIT and MULT are based on the respective best settings used in Table \ref{tab:outofdomain2}.}
We find that the improvements of baseline methods tend to be smaller when the target training set becomes larger.
This is partly because INIT and MULT do not explicitly preserve the parameters from source models in the constructed target models.
Thus, the transferred information is diluted while we train the target model with more data. 
In contrast, our transfer method explicitly saves the transferred information in the base part of our target model, and we use separate learning rates to help the target model to preserve the transferred knowledge.  
Similar experiments on other datasets are shown in the supplementary material.

\subsection{Effect of the Hyper-parameter $\psi$ } 
\begin{figure}[]
	\resizebox{\columnwidth}{!}{\input{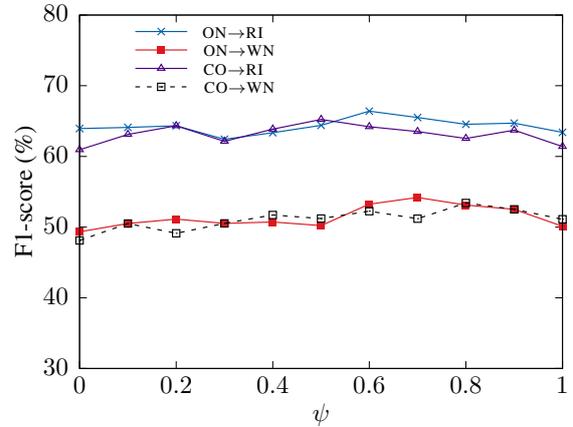}}\vspace{-10pt} 
	\caption{{ 	Performance at different $\psi$ of our methods.}}
	\label{fig:f3}
		\vspace{-10pt}
\end{figure}

We present a set of experiments around the hyper-parameter $\psi$ in~\figref{fig:f3}.
Such experiments over different datasets can shed some light on how to select this hyper-parameter.
{We find that when the target data set is  small (Ritter11), the best $\psi$ are 0.5 and 0.6 respectively for the two source domains, whereas when the target data set is larger (WNUT16), 
the best $\psi$ becomes 0.7 and 0.8. 
The results suggest that the optimal $\psi$ tends to be relatively larger when the target data set is larger.}

\section{Related Work}\label{sec:related}
\vspace{-5pt}

Domain adaptation and transfer learning has been a popular topic that has been extensively studied in the past few years {\cite{DBLP:journals/tkde/PanY10}}.
For well-studied conventional feature-based models in NLP, there are various classic transfer approaches, such as EasyAdapt~\cite{Daum2007FrustratinglyED}, instance weighting~\cite{Jiang2007InstanceWF} and structural correspondence learning~\cite{blitzer2006domain}. 
Fewer works have been focused on transfer approaches for neural models in NLP.
\citet{Mou2016HowTA} use intuitive transfer methods (INIT and MULT) to study the transferability of neural network models for the sentence (pair) classification problem;
{\citet{lee2017transfer} utilize the INIT method on highly related datasets of electronic health records to study their specific de-identification problem.}
\citet{Yang2016TransferLF} use the MULT approach in sequence tagging tasks including named entity recognition. 
Following the MULT scheme, \citet{Wang2018LabelawareDT} introduce a label-aware mechanism into maximum mean discrepancy (MMD) to explicitly reduce domain shift between the same labels across domains in medical data.
Their approach requires the  output space to be the same in both source and target domains.
Note that the scenario in our paper is that the output spaces are different in two domains. 


All these existing works do not use domain-specific embeddings for different domains and they use the same neural model for source and target models.  
However, with our word adaptation layer, it opens the opportunity to use domain-specific embeddings.
Our approach also addresses the domain shift problem at both input and output level by re-constructing target models with our specifically designed adaptation layers.
{The hyper-parameter $\psi$ in our proposed methods and $\lambda$ in MULT both control the knowledge transfer from source domain in the transfer learning process.
	While our method works on top of an existing pre-trained source model directly, MULT needs re-training with source domain data each time they train a target model.
}

{\citet{DBLP:conf/acl/FangC17} add an ``augmented layer'' before their final prediction layer for cross-lingual POS tagging --- which is a simple multilayer perceptron performing local adaptation for each token separately --- {ignoring contextual information}. 
In contrast, we employ a BLSTM layer due to its ability in capturing contextual information, which was recently shown to be crucial for sequence labeling tasks such as NER \cite{DBLP:conf/acl/MaH16,DBLP:conf/naacl/LampleBSKD16}.
We also notice that a similar idea to ours has been used in the recently proposed Deliberation Network~\cite{NIPS2017_6775} for the sequence generation task, where a second-pass decoder is added to a first-pass decoder to polish sequences generated by the latter.
}

We propose to learn the word adaptation layer in our task inspired by two prior studies.
\citet{DBLP:conf/acl/FangC17} use the cross-lingual word embeddings to obtain distant supervision for target languages.
\citet{yang2017} propose to re-train word embeddings on target domain by using regularization terms based on the source-domain embeddings, where some hyper-parameter tuning based on down-stream tasks is required.
{Our word adaptation layer serves as a linear-transformation~\cite{Mikolov2013ExploitingSA}, which is learned based on corpus level statistics.}
{Although there are alternative methods that also learn a mapping between embeddings learned from different domains~\cite{Faruqui2014ImprovingVS,Artetxe2016LearningPB,Smith2017OfflineBW},
such methods usually involve modifying source domain embeddings, and thus re-training of the source model based on the modified source embeddings would be required for the subsequent transfer process.}

\section{Conclusion}\label{sec:conclusion}
We proposed a novel, lightweight transfer learning approach for cross-domain NER with neural networks.
Our introduced transfer method performs adaptation across  two domains using adaptation layers augmented on top of the existing neural model.
Through extensive experiments, we demonstrated the effectiveness of our approach, reporting better results over existing transfer methods.
Our approach is general, which can be potentially applied to other cross-domain structured prediction tasks.
Future directions include investigations on employing alternative neural architectures such as convolutional neural networks (CNNs) as adaptation layers, as well as on how to learn the optimal value for $\psi$ from the data automatically rather than regarding it as a hyper-parameter.
\footnote{We make our supplementary material and code available at  \url{http://statnlp.org/research/ie}.}

\section*{Acknowledgments}

We would like to thank the anonymous reviewers for their thoughtful and constructive comments. 
Most of this work was done when the first author was visiting Singapore University of Technology and Design (SUTD).
This work is supported by DSO grant DSOCL17061, and is partially supported by Singapore Ministry of Education Academic Research Fund (AcRF) Tier 1 SUTDT12015008.

\bibliography{emnlp2018}
\bibliographystyle{acl_natbib_nourl}

\end{document}